%% file: main.tex
\documentclass[conference]{IEEEtran}
\IEEEoverridecommandlockouts
\usepackage{cite}
\usepackage{amsmath,amssymb,amsfonts}
\usepackage[ruled,vlined,linesnumbered]{algorithm2e}
\usepackage{graphicx}
\usepackage{textcomp}
\usepackage{xcolor}
\def\BibTeX{{\rm B\kern-.05em{\sc i\kern-.025em b}\kern-.08em
    T\kern-.1667em\lower.7ex\hbox{E}\kern-.125emX}}

\usepackage{multirow}
\usepackage{makecell}
\usepackage{booktabs}

\usepackage{tikz}
\usetikzlibrary{patterns}
\usepackage{pgfplots}
\pgfplotsset{compat=1.18}
\usetikzlibrary{external}

\usepackage[caption=false]{subfig}
\usepackage[nolist]{acronym}
\usepackage{balance}

\newcolumntype{L}[1]{>{\raggedright\let\newline\\\arraybackslash\hspace{0pt}}m{#1}}
\newcolumntype{C}[1]{>{\centering\let\newline\\\arraybackslash\hspace{0pt}}m{#1}}
\newcolumntype{R}[1]{>{\raggedleft\let\newline\\\arraybackslash\hspace{0pt}}m{#1}}

\begin{document}

\begin{acronym}[TDMA]
    \acro{CNN}{Convolutional Neural Network}
    \acro{MAE}{Mean Absolute Error}
    \acro{RMSE}{Root Mean Squared Error}
    \acro{EER}{Equal Error Rate}
    \acro{ROC}{Receiver Operating Characteristic Curve}
\end{acronym}

\title{Using Deep Neural Networks to Quantify Parking Dwell Time\\
\thanks{This work has been supported by the Brazilian National Council for Scientific and Technological Development (CNPq) -- Grant 405511/2022-1, and by the Coordenação de Aperfeiçoamento de Pessoal de Nível Superior – Brasil (CAPES) – Finance Code 001.}
}

\author{
    \IEEEauthorblockN{Marcelo Marques Ribas\IEEEauthorrefmark{1}, 
        Heloisa Benedet Mendes\IEEEauthorrefmark{1},
    Luiz Eduardo de Oliveira\IEEEauthorrefmark{1},\\
    Luiz A. Zanlorensi\IEEEauthorrefmark{2},
    Paulo Lisboa de Almeida\IEEEauthorrefmark{1}}
    \IEEEauthorblockA{\IEEEauthorrefmark{1}Departamento de Informática (DInf), Universidade Federal do Paraná, Curitiba, PR - Brazil\\\{marcelomarques,heloisamendes,luiz.oliveira,paulorla\}@ufpr.br
    }
    \IEEEauthorblockA{\IEEEauthorrefmark{2}DeepNeuronic, Covilhã - Portugal\\
	   luiz.zanlorensi@deepneuronic.com
    }
}

\maketitle

\begin{abstract}
In smart cities, it is common practice to define a maximum length of stay for a given parking space to increase the space's rotativity and discourage the usage of individual transportation solutions. However, automatically determining individual car dwell times from images faces challenges, such as images collected from low-resolution cameras, lighting variations, and weather effects. In this work, we propose a method that combines two deep neural networks to compute the dwell time of each car in a parking lot. The proposed method first defines the parking space status between occupied and empty using a deep classification network. Then, it uses a Siamese network to check if the parked car is the same as the previous image. Using an experimental protocol that focuses on a cross-dataset scenario, we show that if a perfect classifier is used, the proposed system generates 75\% of perfect dwell time predictions, where the predicted value matched exactly the time the car stayed parked. Nevertheless, our experiments show a drop in prediction quality when a real-world classifier is used to predict the parking space statuses, reaching  49\% of perfect predictions, showing that the proposed Siamese network is promising but impacted by the quality of the classifier used at the beginning of the pipeline.
\end{abstract}


\input{intro}
\input{relatedWorks}
\input{proposal}

\input{protocol}
\input{experiments}
\input{conclusion}

\bibliographystyle{IEEEtran}
\bibliography{IEEEabrv,biblio}

\end{document}

%% file: intro.tex
\section{Introduction}

Developments regarding parking space monitoring through images have been proposed in the recent past, including the development of datasets to be used as benchmarks \cite{almeidaEtAl2015,amatoEtAl2017,nietoEtAl2019}, deep learning-based approaches to classify the individual parking spaces between empty/occupied \cite{amatoEtAl2017,nietoEtAl2019,dhuriEtAL2021}, and the automatic segmentation of the parking spaces positions \cite{almeidaEtAl2023,GrbicKoch2023}.

Following the thread of these deep learning-based innovations, this paper proposes an approach to counting the time a car stays parked in a parking space. To accomplish this, we use images from cameras and process these images using siamese networks. By collecting data about the time each car stays parked, we may generate helpful information about, for instance, the usage of cities' parking areas.

Furthermore, it is a common practice to have parking spaces with a maximum length of stay (e.g., 15 minutes), and a system that counts the time a driver stays parked may help authorities detect offenders. Other usages of such a system may include detecting abandoned or broken cars (e.g., a car parked for too long may be abandoned or broken) and the detection of illegally parked cars (e.g., a car that stayed stationary for too long in an area that is not defined as a parking space).

\begin{figure}[!htbp]
    \centering    
    \subfloat[13:10]{\includegraphics[height=1.5cm]{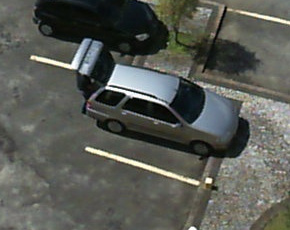}}
    \hspace{0.1cm}
    \subfloat[14:55]{\includegraphics[height=1.5cm]{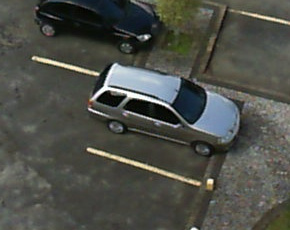}}
    \hspace{0.1cm}
    \subfloat[16:20]{\includegraphics[height=1.5cm]{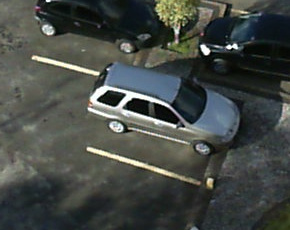}}
    \hspace{0.1cm}
    \subfloat[17:40]{\includegraphics[height=1.5cm]{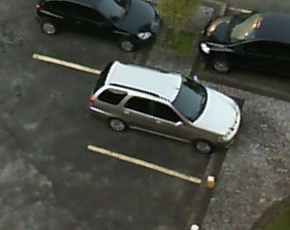}}
    \hspace{0.1cm}
  \caption{A car parked in the same position at different times -- PKLot dataset.}
  \label{fig:sameCarExample}
\end{figure}

\begin{figure}[!htbp]
    \centering
    \subfloat[PKLot]{\includegraphics[height=2.4cm, trim=2cm 0cm 6cm 4cm, clip=true]{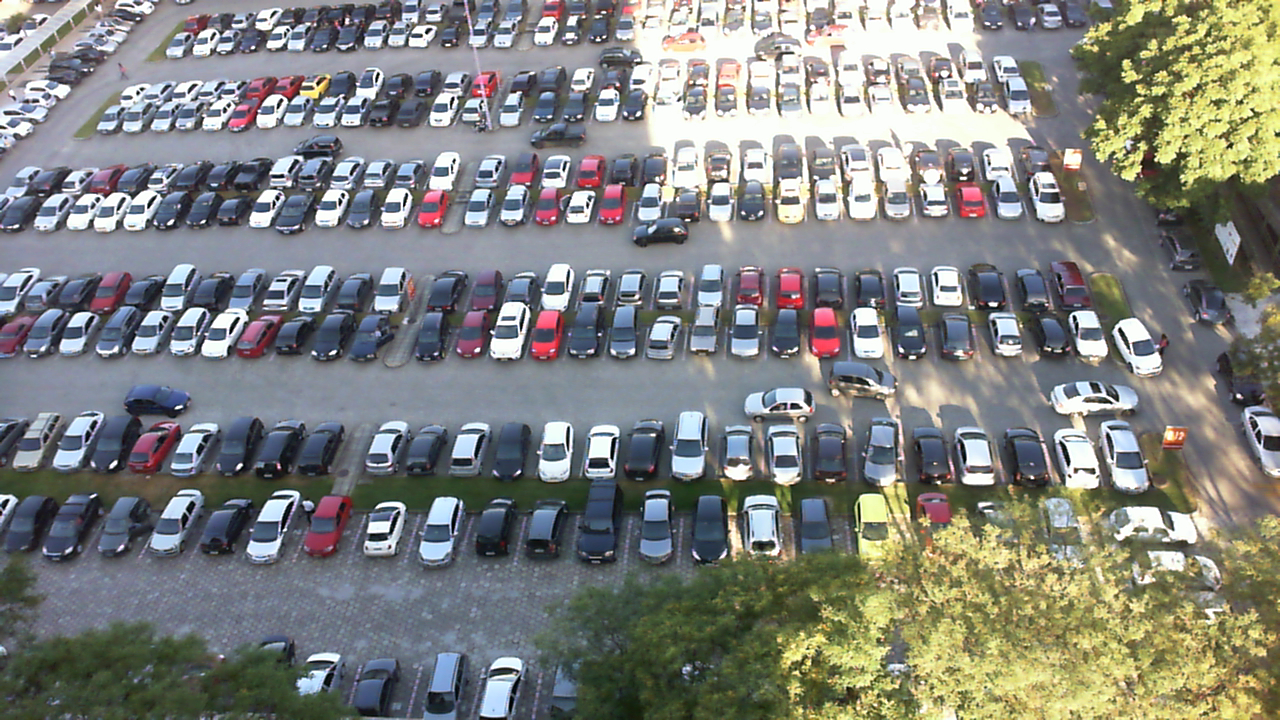}}
    \hspace{0.2cm}
    \subfloat[CNRPark-EXT]{\includegraphics[height=2.4cm,  trim=0cm 0cm 0cm 5cm, clip=true]{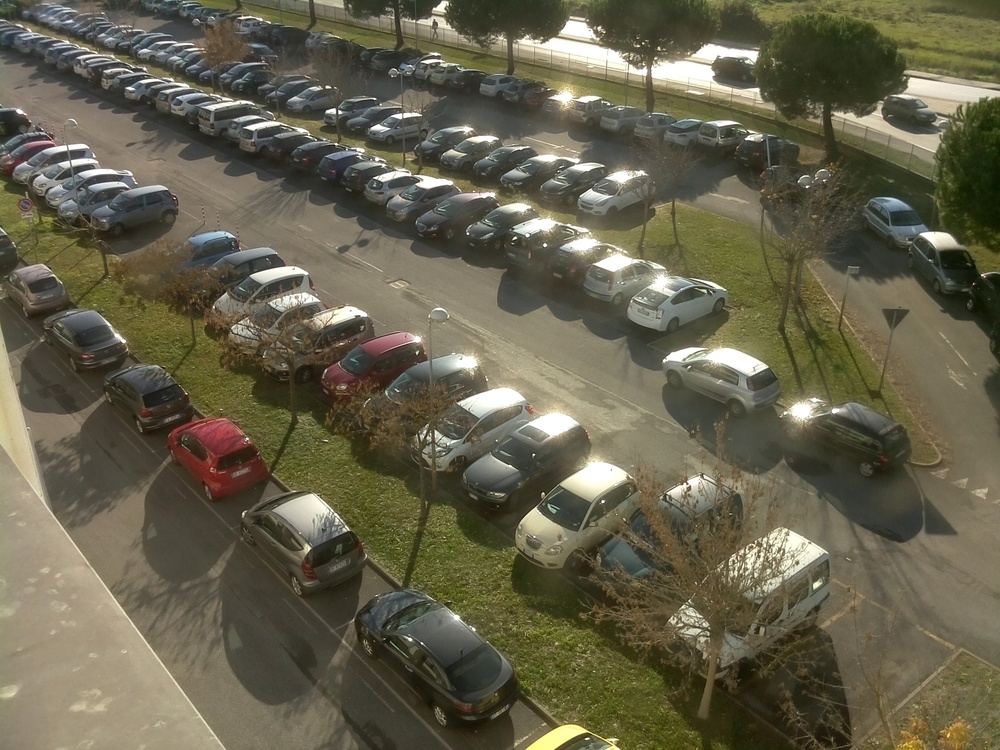}}
  \caption{Image examples from the PKLot and CNRPark-EXT datasets.}
  \label{fig:pklotAntCNR}
\end{figure}

Quantifying the parking time is a complex problem since 1~--~ the vehicles' images are often collected at a distance, with low-resolution cameras (e.g., in the PKLot dataset, the bounding box of a car is 56 x 51 pixels in size, on average); 2~--~vehicles may park in several positions relative to the camera; 3~--~images are often collected with a low frame rate, where many minutes may pass between consecutive images collection (e.g., in the PKLot dataset one image is taken for every 5 minutes); 4~--~changes in luminosity, occlusions, and shadows can make it difficult to compare the vehicles between different images. See an example of the same car parked in the same position at different times in Figure \ref{fig:sameCarExample}. Figure \ref{fig:pklotAntCNR} shows a variety of car angles and luminosity differences in a parking lot.

In this work, we focus on the following contributions:

\begin{itemize}
    \item We define a siamese network to compare cars.

    \item We define a complete pipeline to check the parking status and then verify the dwell time of each car.
\end{itemize}

The complete pipeline includes a classification network to classify the parking spaces, a \ac{CNN}-based siamese network to compare cars, and an algorithm that combines the information of both networks to update the dwell time of each parked car.

The remainder of this work is structured as follows: We show the related works in Section \ref{sec:related}. The proposed approach, including the description of the classification and siamese networks used and the algorithm used to compute the car's dwell time is given in Section \ref{sec:proposed}. In Section \ref{sec:protocol}, we define the experimental protocol, where we define a cross-dataset scenario where no train samples from the target parking lot are given for the system. Section \ref{sec:experiments} shows that the proposed method can reach an \ac{MAE} of 46 and 54 minutes in the PKLot and CNRPark-EXT datasets, respectively. Finally, in Section \ref{sec:conclusion}, we present our conclusions.

%% file: relatedWorks.tex
\section{Related Works}\label{sec:related}

Calculating the duration a car remains parked in a specific location requires using two computer vision tasks: detecting and tracking objects. Deep learning-based object detection stands out as one of the most widely adopted tasks in computer vision. It can be categorized into two types: single-stage and two-stage methods. Single-stage methods, exemplified by YOLO \cite{Redmon16}, abstain from the use of a Region Proposal Network (RPN)\cite{Ren2015} for selecting regions of interest (ROI). Consequently, they demonstrate higher speed, with object locations generated by a single CNN network. In contrast, two-stage methods like Faster R-CNN \cite{Ren2015} and Mask R-CNN \cite{He2017} rely on an RPN network for ROI generation, typically yielding more precise bounding boxes.

Deep learning techniques are also often used in tracking tasks, broadly divided into two categories -- Single-Object Tracking (SOT) and Multiple-Object Tracking (MOT). The key distinction lies in that, in MOT, multiple objects are present within the target scene. Consequently, the model must address various complexities, such as object occlusion and the presence of objects with similar appearances.

The evolution of object tracking can be delineated through three stages. The initial stage, which occurred around the year 2000, primarily witnessed the application of classic algorithms and machine learning in target tracking. While these algorithms boasted features like low computational complexity, swift execution, and minimal hardware requirements, their robustness and accuracy were relatively modest.

Between 2010 and 2016, the second phase of object tracking development saw the rise of methods like MOOSE \cite{Bolme2010} and SORT \cite{Bewley2016}, drawing substantial attention. This prompted researchers to delve into trackers based on correlation filtering, revealing notable advantages in speed and accuracy across various evaluation datasets. SORT has undergone further enhancements in DeepSort \cite{Wojke17} and OCSORT \cite{Cao23}. Recently, Sharma et al. \cite{Sharma23} utilized a combination of these two methods along with YoloV8 to create a real-time algorithm for tracking parking time violations using closed-circuit cameras. 

The ongoing third stage, from 2016 to the present, is characterized by advancements in target tracking driven by deep learning algorithms, particularly the Siamese network. Continual enhancements in robustness and accuracy have been achieved, facilitated by the utilization of increasingly abundant datasets. This progress underscores the formidable end-to-end learning capabilities of deep learning in the realm of object tracking. A comprehensive survey of deep learning for visual tracking can be found in \cite{Zadeh2021}.

%% file: proposal.tex
\section{Proposed Approach}\label{sec:proposed}

We propose using two deep neural networks in a pipeline. First, a classification network is used to classify the parking spaces between occupied and empty. The second network is a siamese comparison network, which will be used to compare cars. For both the classification and siamese Networks, we employ the large version of the MobileNetV3 \cite{howardEtAl2019Mobilenet} as the backbone (about 5M parameters) due to its tradeoff between accuracy and computational cost \cite{HochuliEtAl2023}.

We defined the input size of the images for both networks as $128 \times 128 \times 3$.
To compute the similarity between the input images pair for the siamese network, we use the Contrastive Loss~\cite{chopra2005contrastive,hadsell2006contrastive}. The classification network uses the Cross-entropy loss function. Since the parking space positions are fixed, we use the strategy described in \cite{almeidaEtAl2015}, where for each image, each parking space position is cropped as a rotated rectangle. This cropped image is fed to the networks to define its status (empty/occupied) and to check if the parked car is the same between two consecutive images.

\begin{figure}[!htbp]
    \centering    
    \subfloat[Negative]{\includegraphics[height=1.5cm]{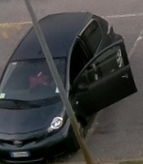}
    \includegraphics[height=1.5cm, trim=0cm 1.5cm 0cm 0cm, clip=true]{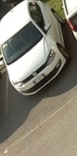}}
    \hspace{1.0cm}
    \subfloat[Positive]{\includegraphics[height=1.5cm]{imgs/TrainSamplesExample/437708_0_camera6.png}
   \includegraphics[height=1.5cm]{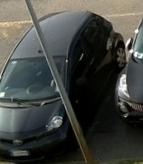}}
    \hfill
  \caption{Siamese network training samples examples.}
  \label{fig:tripletExample}
\end{figure}

The cropped training images and their statuses are used to train the classification network. To train the siamese networks, we consider each individual car in the training set to create the training pairs.
To generate a negative pair, where the images are from different cars, we combine the image of the current car with a random image from another car.
To generate a positive pair, where both images belong to the same car, we combine the car's image with an image of the same car taken at a different random time.
Examples of negative and positive training samples are given in Figure \ref{fig:tripletExample}.

The proposed pipeline is straightforward.
As we are interested in counting the time a car stays parked, we must consider the time between acquiring consecutive images. Considering that one image is taken for every $k$ seconds, and the image of a parking space collected at the current time $t$, we first check if the parking space is occupied or empty using the classification network. If the parking space is occupied, we check the status of the parking space in the previous image, collected at $t-k$.
If the parking space was empty in the previous image, the detected car is considered a new one (the car that has just parked).
Otherwise, the image of the parking space (occupied by a car) collected at time $t$ is compared with the image of the same parking space collected at $t-k$ using the comparison (siamese) network. If the network deems the car to be the same, the time counter for the car is increased by $k$. Otherwise, the car is considered as a new one.

\begin{figure}[htbp]
    \centering
    \includegraphics[height=5.0cm]{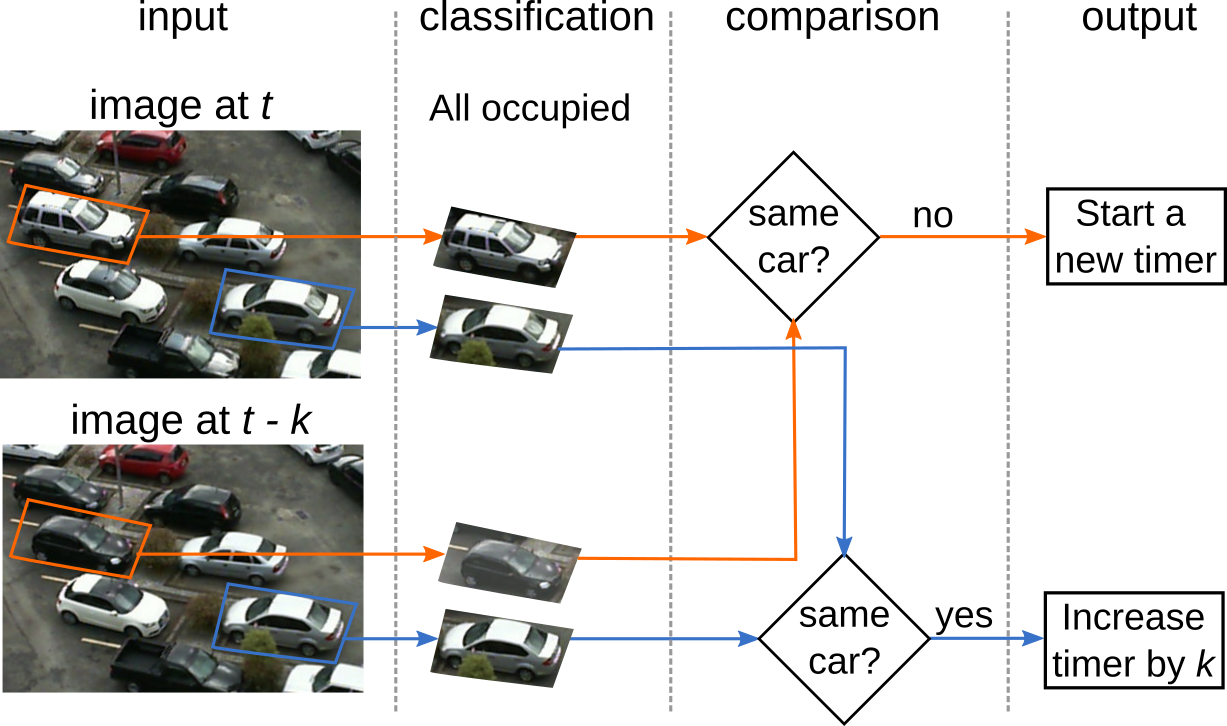}
    \caption{Flowchart example for two parking spaces. Both parking spaces were occupied between $t-k$ and $t$. The parked car in the orange parking space is not the same between images, and a new timer is created. The car in the blue parking space remained the same, and its timer was increased.}
    \label{fig:flowchart}
\end{figure}

Figure \ref{fig:flowchart} illustrates this pipeline for two occupied parking spaces between times $t-k$ and $t$. The complete procedure is depicted in Algorithm \ref{alg:dwellTime}. The algorithm assumes that the classification network classified the images between empty or occupied.

\begin{algorithm}[htpb]
	\footnotesize
	\SetKwInput{KwData}{Input}
	\KwData{$p$: parking space information at the current time $t$; $q$: parking space information at the previous time $t - k$; $k$: the number of seconds between $p$ and $q$; $N$: a trained network to compare cars.}
	\KwResult{The car dwell time is updated}
	\If{$p$ is empty}{
		\Return
	}
    \If{$q$ is empty}{
        $p.time$ = 0\tcp{Car just parked}
		\Return
	}
    $result$ = compare(N, p.image, q.image)\\
    \If{$result$ is same car}{
        $p.time$ = $q.time + k$
    }\Else{
        $p.time$ = 0
    }
    
	\caption{\textsc{dwelltime}$(p,q,k,N)$.}
	\label{alg:dwellTime}
\end{algorithm}

%% file: protocol.tex
\section{Experimental Protocol}\label{sec:protocol}

\subsection{Datasets and Metrics}

We use the well-known PKLot \cite{almeidaEtAl2015} and CNRPark-EXT \cite{amatoEtAl2017} datasets for the experiments. These datasets assume that the camera is at a fixed position and that the parking space positions do not change over time. Image examples from the PKLot and CNRPark-EXT datasets can be seen in Figure \ref{fig:pklotAntCNR}. The time $k$ taken between consecutive images is 5 minutes for the PKLot, whereas for the CNRPark-EXT, $k$ is 30 minutes. A value of 30 minutes between images may be considered too long  (e.g., a car may park for 20 minutes and leave without the system noticing its existence). However, we  use the CNRPark-EXT for its robustness and to include two different datasets in the tests.

To create a ground truth for the experiments, we manually annotated each car with an individual identifier to track it over multiple images. We labeled 16,694 images from the PKLot and CNRPark-EXT datasets to include these new identifiers, making it possible to track 21,538 individual cars.
During the experiments, the developed approach must count how long a car stays parked in a parking space. To compare the prediction with the ground truth, we use the \acf{MAE} and \acf{RMSE} values, defined as

\begin{equation}
    MAE = \frac{1}{N}\sum_{i=1}^{N}|y_i - \hat{y}_i|
\end{equation}

\begin{equation}
    RMSE = \sqrt{\frac{1}{N}\sum_{i=1}^{N}(y_i - \hat{y}_i)^2}
\end{equation}

\noindent where $N$ is the number of predictions, $y_i$ is the actual amount of time the car stayed parked according to the ground truth, and $\hat{y}_i$ is the predicted time.

When a single car is deemed as a new one multiple times, we use its first prediction $\hat{y}$ to compare with the ground truth $y$, and the remaining predictions are compared to zero (i.e., $0 - \hat{y}$ in the equations). A similar approach is used when a car is not detected, where the ground truth will be compared with zero, as the system computed no time for the car. We also show the accuracy achieved, defined as the number of cars where the predicted parking time matched exactly with the ground-truth, divided by the total number of parked cars in the dataset.

\subsection{Networks and Training Procedure}

Both the classification and siamese (comparison) deep networks were pre-trained in the ImageNet dataset. We employed the Adam optimizer with an initial learning rate of $0.001$ and a batch size of 32 for both networks. As suggested in \cite{Almeida2022}, both the classification and siamese networks are trained in a cross-dataset fashion. This is important since we are putting the system to the test without considering training samples from the target parking lot, creating a realistic scenario where the system can be deployed without training it in a target scenario.
First, we trained the networks using images from the PKLot dataset and tested them in the CNRPark-EXT dataset. Then, we reversed the training and test sets, training using the CNRPark-EXT images and testing using the PKLot images.

The first 70\% of day images of each camera from the training dataset are used for training, and the remaining 30\% for validation.
Table \ref{tab:splits} shows the training, validation, and testing splits\footnote{The number of samples available in each dataset can be different from the original datasets~\cite{almeidaEtAl2015}, \cite{amatoEtAl2017} since we manually labeled new samples.}. For each epoch, the complete training set is used to train the classification network. Considering the siamese comparison network, since there is a quadratic number of possible training and validation pairs relative to the number of available cars in the datasets, we select 20,000 random training pairs and 20,000 random validation pairs for each epoch.

\begin{table}[htpb]
\centering
\caption{Training, validation, and testing splits.}
\label{tab:splits}
\footnotesize
\begin{tabular}{lll}
\toprule
\multicolumn{3}{l}{Training using PKLot test using CNRPark-EXT}\\\midrule
\#Training & \#Validation & \#Test \\\cmidrule(lr){1-1}\cmidrule(lr){2-2}\cmidrule(lr){3-3}
\makecell[l]{309,381 occupied\\360,459 empty}&\makecell[l]{143,529 occupied\\113,754 empty}& \makecell[l]{83,655 occupied\\67,972 empty}\\\midrule

\multicolumn{3}{l}{Training using CNRPark-EXT test using PKLot}\\\midrule
\#Training & \#Validation & \#Test \\\cmidrule(lr){1-1}\cmidrule(lr){2-2}\cmidrule(lr){3-3}
\makecell[l]{58,625 occupied\\52,958 empty}&\makecell[l]{25,030 occupied\\15,014 empty}&\makecell[l]{452,730 occupied\\474,213 empty}\\\bottomrule
\end{tabular}
\end{table}

The classification and siamese networks are fine-tuned for 30 and 100 epochs, respectively. We select the model with the lowest loss in the validation set as the final model\footnote{We also defined that the siamese networks must be trained for at least 15 epochs since the training/validation samples are taken at random for each epoch, which can create loss instabilities.}.
For the classification network, for each epoch, we define a 10\% chance of applying one of the following data augmentation techniques in each training image: Horizontal Flip,
Gaussian Blur with a $5\times 9$ kernel and standard deviation $\in[0.1, 5]$, Random Crop, and Random Autocontrast. For the siamese network, we define a 10\% chance of applying a brightness change $\in[-0.2, 0.2]$ for each training pair.

We use the validation sets for both the classification and comparison networks to define the classification threshold. We defined the optimal threshold as the one that generates the \ac{EER} point in the \ac{ROC} for the classification network. For the siamese Networks, we defined the classification threshold as the biggest threshold value with at most a 5\% chance of mistaking a different pair of cars as a pair of the same car.

%% file: experiments.tex
\section{Experiments}\label{sec:experiments}

In Section \ref{subsec:timeUsingGroundTruth}, we begin presenting the results considering a perfect parking spaces classifier (that predicts if a parking space is empty or occupied) to check the ability of the comparison (siamese) network to track cars over multiple images. Then, in Section \ref{subsec:timeUsingClassifier}, we present the results considering the complete pipeline, considering a classification network followed by a comparison network, showing the expected behavior of the proposed approach in a real-world scenario. All results presented are an average of 5 runs.

\subsection{Measuring the Dwell Time using the Ground Truth} \label{subsec:timeUsingGroundTruth}

For the first set of experiments, we want to analyze the ability of the trained siamese network to correctly track the presence of the same car over multiple images of a parking space. To accomplish this, we check if the parking space is occupied or empty using the information available in the ground truth since we are interested in checking the performance of the siamese network without the interference of other factors.

First, to measure the competence of the siamese networks to compare cars, for each car\footnote{The same car may appear over multiple images. We randomly selected just one image of the car.}
in the testing set, we randomly selected an image of the same car (in a different moment) to create a positive pair, and a random image from a different car to generate a negative pair. Thus, in a test dataset containing $n$ individual cars, we generated $2n$ testing pairs. The results of the trained networks in these testing images are shown in Table \ref{tab:resultsSiamese}.

{
\begin{table}[htb]
\centering
\setlength{\tabcolsep}{0.4em}
\caption{Siamese network results.}
\label{tab:resultsSiamese}
\begin{tabular}{@{}llrr@{}}
\toprule
Training Set & Testing Set & \# Test Pairs & \multicolumn{1}{c}{Accuracy (stdev)} \\\midrule
PKLot & CNPark-EXT & 15,482 & 93.8\% (1.6) \\
CNPark-EXT & PKLot & 27,594 & 96.2\% (1.8)\\\midrule
\textbf{Total} & & 43,076 & 95.3$\%^*$ \\\bottomrule
\multicolumn{4}{l}{\scriptsize{*Weighted average.}}\\
\end{tabular}
\end{table}
}



The results in Table \ref{tab:resultsSiamese} show that the networks achieved good generalizations, with accuracies of 95.3\%, on average. This is an interesting result since, despite no image samples from the target parking lot being given for training, most of the car pairs were correctly recognized by the networks.

In Table \ref{tab:metricsUsingGT} we show the results achieved by using the trained siamese networks to define the parking dwell time of each car in the test sets, using the approach discussed in Section \ref{sec:proposed}. As one can observe, the proposed approach achieved promising results, reaching perfect predictions for 69.1\% and 77.8\% in the PKLot and CNRPark-EXT tests, respectively. The \ac{MAE} and \ac{RMSE} results are shown in minutes. The \ac{MAE} values show that, on average, the proposed approach misses the correct dwell time with an average error of 40.2 and 47 minutes for the PKLot and CNRPark-EXT datasets, respectively. It is worth remembering that in the CNRPark-EXT dataset, the time difference between two consecutive images is 30 minutes. Thus, one single mistake in detecting the same car in two consecutive images will generate an error of at least 30 minutes.

It is worth noticing that the high number of perfect predictions combined with the relatively high \ac{MAE} suggests that when the system cannot correctly predict the dwell time, the errors generated tend to be large. This result is corroborated by the high \ac{RMSE} values since, due to the quadratic nature of the \ac{RMSE} computation, it penalizes higher differences between the ground truth and the predictions.

\begin{table}[htpb]
\centering
\setlength{\tabcolsep}{2.5pt}
\caption{Results using the ground-truth to get each parking space status (empty or occupied).}
\label{tab:metricsUsingGT}
\begin{tabular}{llrrr}
\hline
Test Set & Camera & MAE (stdev) & RMSE (stdev) & \makecell[c]{\% perfect\\ predictions (stdev)} \\\hline
\multirow{4}{*}{PKLot} &  UFPR04 & 52.9 (1.5) & 111.6 (2.1) & 59.4 (1.0)\\
& UFPR05 & 58.1 (1.7) & 125.1 (1.3) & 58.1 (1.5)\\
& PUC & 29.4 (1.1) & 89.4  (1.9) & 76.3 (0.6)\\\cline{2-5}
& \textbf{Average} & \textbf{40.2 (1.3)} & \textbf{103.0 (1.7)} & \textbf{69.1 (0.8)}\\\hline
\multirow{10}{*}{CNRPark-EXT} & cam1 & 47.8 (11.7) & 117.1 (13.7) & 76.6 (5.9)\\
& cam2 & 33.2 (20.4) & 93.0  (30.3) & 82.7 (9.1)\\
& cam3 & 42.2 (12.7) & 107.9 (13.8) & 78.7 (7.2)\\
& cam4 & 44.4 (11.2) & 114.0 (12.2) & 78.5 (6.1)\\
& cam5 & 43.4 (9.9)  & 114.5 (9.6)  & 80.1 (5.7)\\
& cam6 & 47.5 (9.1)  & 121.5 (8.1)  & 77.8 (5.5)\\
& cam7 & 53.7 (7.5)  & 127.5 (6.9)  & 75.7 (4.5)\\
& cam8 & 44.9 (9.2)  & 113.4 (10.1) & 78.6 (4.9)\\
& cam9 & 57.5 (12.4) & 129.8 (11.2) & 73.5 (6.4)\\\cline{2-5}
& \textbf{average} & \textbf{47.0 (10.4)} & \textbf{117.8 (10.8)} & \textbf{77.8 (5.8)}\\\hline
\end{tabular}
\end{table}



\subsection{Measuring the Dwell Time using a Classifier} \label{subsec:timeUsingClassifier}

In this Section, we show the results of the complete proposed approach (Section \ref{sec:proposed}), where a network is first used to classify the parking spaces between occupied and empty, and the siamese network is used to compare the cars. This test was developed to answer how a classifier that predicts the parking space status (and may incorrectly classify it) can hinder vehicle dwell time accounting.

First, in Table \ref{tab:resultsClassification}, we show the averaged classification results of the classification networks when put to the test in the target (testing set) parking lot to classify its parking spaces between occupied and empty. As one can observe, we reached an average classification accuracy of 92.4\%, which is similar to the results achieved by most of the methods in the state-of-the-art when we consider a cross-dataset scenario \cite{Almeida2022,HochuliEtAl2023}.

{
\begin{table}[htbp]
\centering
\setlength{\tabcolsep}{0.4em}
\caption{Classification network results.}
\label{tab:resultsClassification}
\begin{tabular}{@{}llrr@{}}
\toprule
Training Set & Testing Set & \# Test Samples & \multicolumn{1}{c}{Accuracy (stdev)} \\\midrule
PKLot & CNPark-EXT & 160,298 & 93.0\% (0.24)\\
CNPark-EXT & PKLot & 926,943 & 92.3\% (0.15)\\\midrule
\textbf{Total} & & 1,087,241 & 92.4\%$^*$\\\bottomrule
\multicolumn{4}{l}{\scriptsize{*Weighted average.}}\\
\end{tabular}
\end{table}
}

In Table \ref{tab:metricsUsingPipeline}, we show the results of the complete pipeline, using the classification networks to verify if the parking space is occupied or empty, to only then use the siamese networks to check if the car is the same between images and increase the dwell time of the vehicle if necessary. As one can observe, despite the high accuracy of the classification networks, there was a substantial drop in all metrics when compared with the results available in Section \ref{subsec:timeUsingGroundTruth}.

\begin{table}[htpb]
\centering
\setlength{\tabcolsep}{2.5pt}
\caption{Results using the complete pipeline.}
\label{tab:metricsUsingPipeline}
\begin{tabular}{llrrr}
\hline
Test Set & Camera & MAE (stdev) & RMSE (stdev) & \makecell[c]{\% perfect\\ predictions (stdev)} \\\hline
\multirow{4}{*}{PKLot} & UFPR04 & 68.7 (5.5) & 141.8 (9.2) & 21.7 (5.4)\\
& UFPR05 & 75.2 (13.9) & 151.6 (21.5) & 18.3 (5.5)\\
& PUC & 34.5 (2.6) & 96.2 (5.9) & 56.9 (6.6)\\\cline{2-5}
& \textbf{Average} & \textbf{53.3 (5.4)} & \textbf{124.6 (9.2)} & \textbf{38.4 (4.8)}\\\hline
\multirow{10}{*}{CNRPark-EXT} & cam1 & 107.9 (6.4) & 175.2 (6.5) & 43.8 (2.2)\\
& cam2 & 46.2 (3.9) & 114.2 (5.6) & 75.6 (2.3)\\
& cam3 & 82.8 (3.9) & 158.2 (3.5) & 57.3 (2.5)\\
& cam4 & 66.5 (3.5) & 136.5 (4.4) & 61.5 (1.1)\\
& cam5 & 86.2 (4.2) & 156.2 (4.9) & 52.0 (1.5)\\
& cam6 & 80.2 (3.6) & 151.6 (4.1) & 56.8 (0.9)\\
& cam7 & 101.6 (5.0) & 170.0 (4.5) & 46.5 (2.6)\\
& cam8 & 74.4 (1.7) & 144.0 (2.4) & 58.8 (0.7)\\
& cam9 & 88.4 (5.7) & 156.9 (7.4) & 51.9 (1.7)\\\cline{2-5}
& \textbf{average} & \textbf{84.5 (1.4)} & \textbf{155.0 (1.8)} & \textbf{54.3 (0.7)}\\\hline
\end{tabular}
\end{table}

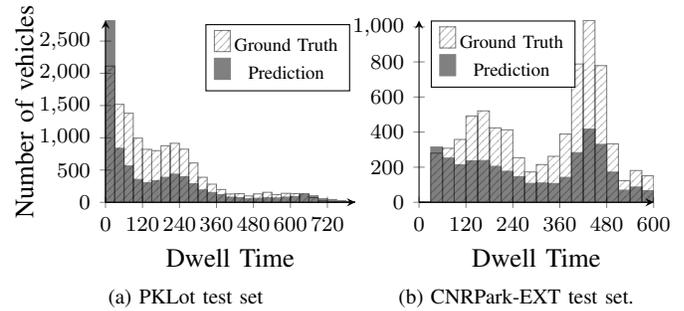
\begin{figure}[htbp]
    \centering
    \hspace{-0.4cm}
    \subfloat[PKLot test set]{\input{./tikz/histPKLot.tikz}}
    \hspace{-0.4cm}
    \subfloat[CNRPark-EXT test set.]{\input{./tikz/histCNR.tikz}}
    \hfill
  \caption{Car's Dwell Time Histogram using the a) PKLot and b) CNRPark-EXT datasets as test sets.}
  \label{fig:finalTestsHist}
\end{figure}

\begin{figure}[htbp]
    \centering
    \subfloat[07:14]{\includegraphics[height=1.4cm]{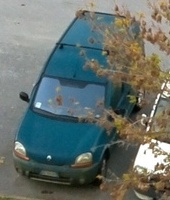}}
    \hspace{0.05cm}
    \subfloat[07:44]{\includegraphics[height=1.4cm]{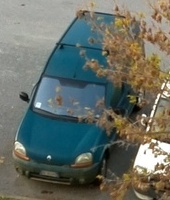}}
    \hspace{0.05cm}
    \subfloat[08:14]{\includegraphics[height=1.4cm]{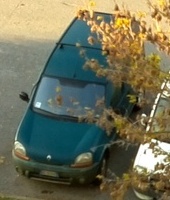}}
    \ \ \dots\ \ 
    \subfloat[13:44]{\includegraphics[height=1.4cm]{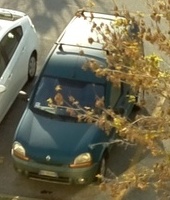}}
    \hspace{0.05cm}
    \subfloat[14:14]{\includegraphics[height=1.4cm]{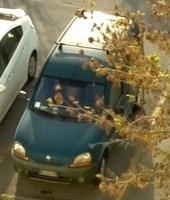}}
    \hspace{0.05cm}
    \subfloat[14:44]{\includegraphics[height=1.4cm]{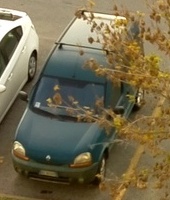}}
    \hspace{0.05cm}
  \caption{A car that stay parked for several hours -- CNRPark-EXT dataset.}
  \label{fig:sameCarCNR}
\end{figure}

This severe drop in the metrics can be explained by the fact that, on average, a car can be parked for several hours in both datasets. On average, the same car is present in 25 (2 hours and 5 minutes) and 12 (6 hours) consecutive images in the PKLot and CNRPark-EXT datasets, respectively. One example is given in Figure \ref{fig:sameCarCNR}. Considering the example, if the classifier incorrectly predicts a single image of the car as a free parking space, the dwell time computation can be severely impacted, especially if the misclassified image is near the beginning of the sequence, since the timer for the current car will be stopped much earlier than it should and a timer may (incorrectly) be started for a new car.

This behavior can be seen in Figure \ref{fig:finalTestsHist}, where we show histograms of the number of cars parked for different time spans considering both the PKLot and CNRPark-EXT datasets. The proposed method overestimated the number of cars with a low stay times, between 0 and 30 minutes, for both the PKLot and CNRPark-EXT datasets. This can be explained by the car's chronometers being stopped prematurely due to a misclassification of the classification network or an error generated by the siamese network.

\subsection{Lessons Learned} \label{subsec:lessonsLearned}


\textbf{The time span between consecutive images impacts the results: } As discussed in this work, we consider the time span between two consecutive images taken as $k$. Large values of $k$ can lead to more lightweight systems since the number of images taken and processed by minute can be reduced. Nevertheless, large values of $k$ can lead to larger errors since one misclassified image can lead to an error of at least $k$ seconds. Moreover, the greater the value of $k$, the more changes between consecutive images may occur, hindering the ability of the siamese networks to compare cars.

\textbf{The classification network is a key point in generating better results: } Even a classification network with a relatively small error (7.6\% in our experiments) can severely decrease the estimation of the dwell time since a car stays parked for several hours, and one single misclassification can prematurely stop the chronometer for a given car, and start a chronometer for a new car that does not exist in reality. As discussed in \cite{Almeida2022}, a key point that should be tackled is the development of generic classifiers with accuracies near 100\%, which could significantly improve the performance of our system. Classifiers that achieve accuracies of nearly 100\% do exist \cite{almeidaEtAl2015,amatoEtAl2017,Almeida2022}. Nevertheless, such classifiers need training samples from the target parking lot. This work focused on a general approach that does not need training samples from the target parking lot. When training samples from the target parking lot are available, we should expect results near the ones shown in Section \ref{subsec:timeUsingGroundTruth}, where a hypothetical perfect classifier was used.

\textbf{When the system gets things wrong, it really gets it wrong: } The \ac{RMSE} values are much bigger than the \ac{MAE} values in all experiments. This indicates that for some difficult samples, the system can generate big errors, while for most samples, the errors are relatively small (the \ac{RMSE} penalizes bigger errors due to its quadratic nature). These difficult samples should be taken into account in future works.




%% file: tikz/histPKLot.tikz
\begin{tikzpicture}
    \begin{axis}[
        ybar,
        width=4.9cm,
        height=4cm,
        ticklabel style = {font=\footnotesize},
        axis x line=bottom,
        axis y line=left,
        xtick distance=120,
        ytick distance=500,
        scaled y ticks=false,
        xlabel=Dwell Time,
        ylabel=Number of vehicles,
        y label style={at={(-0.25,0.5)}},
         legend style={nodes={scale=0.7, transform shape}}
    ]
        \addplot+ [ybar interval, opacity=0.5, color=black, pattern = north east lines, pattern color = black] table [x=Bin,y=True, col sep=comma]  {tikz/trainCnrTestPklot.csv};
        \addlegendentry{Ground Truth};
        \addplot+ [ybar interval, opacity=0.5, color=black] table [x=Bin,y=Pred, col sep=comma]  {tikz/trainCnrTestPklot.csv};
        \addlegendentry{Prediction};
    \end{axis}
\end{tikzpicture}

%% file: tikz/histCNR.tikz
\begin{tikzpicture}
    \begin{axis}[
        ybar,
        width=4.7cm,
        height=4cm,
        ticklabel style = {font=\footnotesize},
        axis x line=bottom,
        axis y line=left,
        xtick distance=120,
        ytick distance=200,
        scaled y ticks=false,
        legend style={at={(0.05,1.0)},anchor=north west},
        xlabel=Dwell Time,
        legend style={nodes={scale=0.7, transform shape}}
    ]
    \addplot+ [ybar interval, opacity=0.5, color=black, pattern = north east lines, pattern color = black] table [x=Bin,y=True, col sep=comma]  {tikz/trainPklotTestCnr.csv};
    \addlegendentry{Ground Truth};
    \addplot+ [ybar interval, opacity=0.5, color=black] table [x=Bin,y=Pred, col sep=comma]  {tikz/trainPklotTestCnr.csv};
    \addlegendentry{Prediction};
    \end{axis}
\end{tikzpicture}

%% file: conclusion.tex
\section{Conclusion}\label{sec:conclusion}

In this work, we showed a complete pipeline that includes a deep classification network followed by a siamese network to compute the dwell time of vehicles in parking spaces. We evaluated the proposed approach in a cross-dataset scenario, which is the most challenging scenario for vision-based parking spaces monitoring systems \cite{Almeida2022}, where no training samples from the target parking lot are given. To the best of our knowledge, this is the first work that proposes a pipeline to compute the dwell time of cars using images that define a robust protocol, including a cross-dataset scenario and experiments using large volumes of data and the use of well-known metrics, such as \ac{MAE} and \ac{RMSE}.

Our experiments showed promising results for the proposed approach regarding the Siamese networks responsible for comparing cars between images. Our results show that such a system can perfectly compute the dwell time for 69\% and 78\% of the cars for the PKLot and CNRPark-EXT datasets, given that a hypothetical perfect classification network can tell if each parking space is occupied or empty. It is worth mentioning that many classification networks can reach accuracies near 100\%, given that training samples from the target parking lot are given \cite{almeidaEtAl2015,Almeida2022,amatoEtAl2017}.

As we focus on cross-dataset scenarios, we included the experiments regarding the complete pipeline, where a classification network that did not receive training samples from the target parking lot is employed to classify each parking space before computing the dwell time. Our results show that even a network with a relatively small error (7.6\% in our tests) can severely deteriorate the results since the chronometer of the cars can be stopped much earlier than they should. This indicates that, as suggested in \cite{almeidaEtAl2015}, parking space classification networks that do not rely on train instances of the target parking lot should be improved to reach accuracies, at least near to the ones trained in the target parking space.

For future work, we plan to improve the training of both the siamese and classification networks to reduce the error in the cross-dataset scenario and the total error of the complete pipeline, taking into consideration the Algorithm \ref{sec:proposed}.